\documentclass{article}

\usepackage[preprint]{neurips_2022}

\usepackage{subfigure}
\usepackage{multicol}
\usepackage{pdfpages}
\usepackage{multirow}
\usepackage{threeparttable}
\usepackage{amsmath}
\usepackage[utf8]{inputenc} 
\usepackage[T1]{fontenc}    
\usepackage{hyperref}       
\usepackage{url}            
\usepackage{booktabs}       
\usepackage{amsfonts}       
\usepackage{nicefrac}       
\usepackage{microtype}      
\usepackage{xcolor}         
\usepackage{natbib}
\setcitestyle{numbers,square}

\title{Multi-view Human Body Mesh Translator}

\author{%
  Xiangjian Jiang
  \thanks{Equal contribution} \\
  School of Computer Science and Engineering \\
  Beihang University \\
  China \\
  \And
  Xuecheng Nie $^*$\\
  MT Lab \\
  Meitu Inc. \\
  China \\
  \AND
  Zitian Wang \\
  Institute of Artificial Intelligence \\
  Beihang University \\
  China \\
  \And
  Luoqi Liu \\
  MT Lab \\
  Meitu Inc. \\
  China \\
  \And
  Si Liu \thanks{Corresponding author} \\
  Institute of Artificial Intelligence \\
  Beihang University \\
  China \\
}

\begin{document}

\maketitle
\begin{abstract}

Existing methods for human mesh recovery mainly focus on single-view frameworks, but they often fail to produce accurate results due to the ill-posed setup. Considering the maturity of the multi-view motion capture system, in this paper, we propose to solve the prior ill-posed problem by leveraging multiple images from different views, thus significantly enhancing the quality of recovered meshes. In particular, we present a novel \textbf{M}ulti-view human body \textbf{M}esh \textbf{T}ranslator (MMT) model for estimating human body mesh with the help of vision transformer. Specifically, MMT takes multi-view images as input and translates them to targeted meshes in a single-forward manner. MMT fuses features of different views in both encoding and decoding phases, leading to representations embedded with global information. Additionally, to ensure the tokens are intensively focused on the human pose and shape, MMT conducts cross-view alignment at the feature level by projecting 3D keypoint positions to each view and enforcing their consistency in geometry constraints. Comprehensive experiments demonstrate that MMT outperforms existing single or multi-view models by a large margin for human mesh recovery task, notably, 28.8\% improvement in MPVE over the current state-of-the-art method on the challenging HUMBI dataset. Qualitative evaluation also verifies the effectiveness of MMT in reconstructing high-quality human mesh. Codes will be made available upon acceptance.

\end{abstract}

\section{Introduction}
\label{intro}

Human Mesh Recovery (HMR) from RGB images \cite{kanazawa2018end, kolotouros2019convolutional, georgakis2020hierarchical} is a fundamental task in Computer Vision, aiming to estimate 3D vertices and their topology to model the body shape.
It has a wide range of applications in sports motion analysis \cite{zecha2018kinematic, zhu2020reconstructing, scott20174d}, security surveillance \cite{su2017pose, zheng2019pose} and also plays a crucial role in building the Metaverse \cite{zhang2012microsoft, han2022virtual}.

\begin{figure}[t!]
\centering
\includegraphics[width=\textwidth]{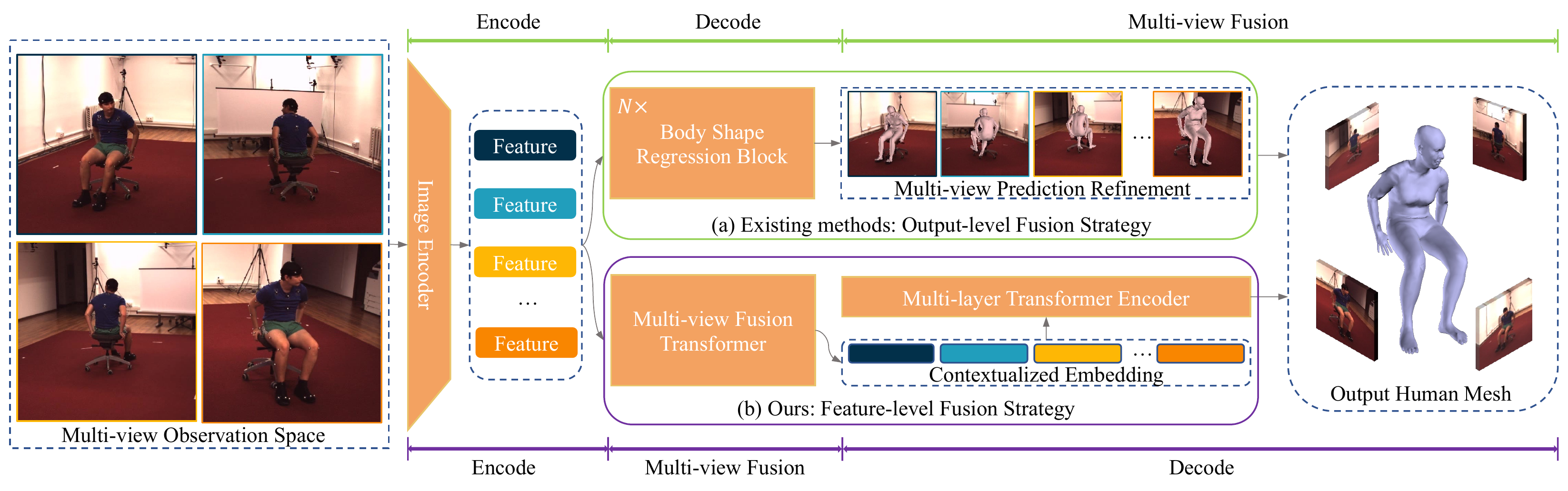}
\vspace{-3mm}
\caption{Comparison between the proposed MMT model and existing ones. After extracting image-level features from different views, (a) Existing models conduct output-level fusion, which fails to capture feature interactions of multiple views; (b) Differently, MMT conducts feature-level fusion, resulting in a way to sufficiently leverage multi-view priors for deriving more accurate estimations. 
}
\vspace{-6mm}
\label{pipeline}
\end{figure}

In literature, existing works mainly follow the path to recovering 3D human mesh from a single monocular RGB image~\cite{kocabas2021pare, sun2021monocular, zanfir2021thundr, kanazawa2018end, moon2020i2l} with neural networks, \emph{e.g.},
Convolutional Neural Networks (CNNs) and Vision Transformer (ViT).
However, they suffer from the severe ill-posed problem due to 
the ambiguity when lifting human bodies from 2D onto 3D space.
Some pioneering attempts~\cite{trumble2018deep, liang2019shape, li20213d, sengupta2021probabilistic, yu2022multi} have tried to bridge the gap by post-processing estimations from multiple views for refining human meshes, as shown in Fig.~\ref{pipeline}(a), but they fail to exploit features of different views directly, resulting in insufficient usage of multi-view priors and achieving uncompetitive performance to single-view counterparts~\cite{zeng20203d, lin2021end, lin2021mesh}. 
Besides, thanks to the growing maturity of the motion capture system, large-scale multi-view datasets for reconstructing volumetric human body representations~\cite{ionescu2013human3, li2019learning, yu2020humbi, peng2021neural} have become available.
Thus, there is an urgent need to develop an effective solution for recovering high-quality 3D human body mesh from multiple camera views.

Motivated by this, we propose learning to fuse view-wise features to produce accurate human body meshes adapted to different viewpoints. Specifically, we present a novel non-parametric model, \textbf{M}ulti-view human body \textbf{M}esh \textbf{T}ranslator (MMT) for this purpose. MMT translates consecutive multi-view images to corresponding targeted meshes, mimicking the language translation process from origin to target. Different from existing methods with output-level fusion, MMT conducts \emph{feature-level fusion}, which fuses multi-view features to contextualized embeddings for decoding mesh vertices of targeted subjects, as shown in Fig.~\ref{pipeline}(b). This encoding-fusing-decoding scheme can sufficiently leverage multi-view priors, overcoming drawbacks of existing encoding-decoding-fusing that neglects considerable feature-level interaction.
In addition, MMT performs estimations on human mesh in a global manner according to multi-view features instead of depending on intermediate and local representations, leading to coherent results for all views. Moreover, MMT introduces cross-view alignment via the fusion of features from multi-view positions mapped to the same 3D keypoints. This feature-level geometry constraint further improves mesh consistency.  

In particular, we implement the proposed MMT model with Vision Transformer (ViT). MMT takes multi-view images as input and outputs the corresponding body meshes for each view. 
Following the convention, MMT first utilizes a CNN backbone to extract high-level features from original images. Then, MMT introduces Multi-view Fusion Transformer to perform feature fusion, which
organizes encoded features as a token sequence and produces context-aware feature embedding based on the underlying interactions among different views.
To align features from different views, MMT conducts 3D human pose estimation as an auxiliary task and projects predicted keypoints into all views to match the corresponding ground truth.
This task guarantees that the fusion tokens are embedded with cross-view consistent semantic clues for human body mesh reconstruction before providing final results.
Finally, with given contextualized embedding, MMT uses another multi-layer transformer encoder with progressive dimensionality reduction \cite{lin2021end} as a decoder to reconstruct the 3D human pose and shape progressively.

Extensive experiments on Human3.6M \cite{ionescu2013human3} and HUMBI \cite{yu2020humbi} benchmarks show the effectiveness of MMT for recovering accurate human body meshes in a multi-view way.
Quantitatively, MMT outperforms current state-of-the-art~\cite{lin2021mesh} by 28.8\% in MPVE on HUMBI.
In addition, solid ablation studies are conducted for each component within MMT and alternative model designs to reach a general conclusion on model performance.

Our contributions are summarized as follows:
1) We propose a novel multi-view model for tackling the human body mesh recovery task. Our model conducts feature-level fusion to sufficiently leverage multi-view priors instead of output-level, leading to notably improved performance.
2) We design a novel cross-view alignment module to fuse the semantic information relevant to human pose and shape from different views with geometry constraints, helping to produce view-wise consistent results.
3) Our model surpasses previous ones under both single-view \cite{lin2021mesh} and multi-view conditions \cite{li20213d} and sets new state-of-the-art.

\section{Related Works}
\label{related}
\paragraph{Human Mesh Recovery} 
Current methods for human mesh recovery can be classified by the number of input views.
The single-view methods estimate the 3D human body shape from a single monocular image \cite{kocabas2021pare, sun2021monocular, zanfir2021thundr, kanazawa2018end, moon2020i2l, choi2020pose2mesh, lin2021end, kolotouros2019convolutional, lin2021mesh, zeng20203d}, and they inevitably fail to make precise predictions with severe ambiguity during the lifting from 2D plane to 3D space.
Therefore, multi-view methods rely on several synchronized camera views to mitigate the ill-posed problem.
Existing works mainly focus on the post-processing and combining the results from parallel single-view models \cite{trumble2018deep, li20213d, li2021detailed, sengupta2021probabilistic, liang2019shape} and even use the temporal dimension \cite{huang2021dynamic} to improve the model performance further.
However, this pipeline ignores underlying interactions of image features from different views and makes the model overdependent on the accuracy of single-view predictions.
Meanwhile, above methods are mostly based on parametric human model, such as SMPL \cite{loper2015smpl}, SMPL-X \cite{pavlakos2019expressive} and STAR \cite{osman2020star}.
Consequently, they usually find it hard to reconstruct the human body of rare distribution due to the limited number of samples used to obtain the parametric models.

Unlike the existing multi-view solutions, our proposed method especially attends to images captured by synchronized cameras and fuses multi-view features. Then the non-parametric model makes final inferences on 3D human pose and shape with transformer processing the vertex-to-vertex relationship.

\paragraph{Vision Transformers}
Impressed by the great success of transformer in Natural Language Processing \cite{devlin2018bert, brown2020language, vaswani2017attention}, there have been pioneering works trying to extend this framework to 3D human pose estimation and mesh recovery \cite{zhang2021direct, lin2021end}.
To the best of our knowledge, existing works on the HMR task have not used transformer to handle the multi-view inputs directly as \cite{zhang2021direct} does in pose regression.
Therefore, we propose a novel method with vision transformer as the core component to process multi-view features and generate fusion tokens closely related to human pose and shape.

\section{Method}
\label{method}
\begin{figure}[t]
\centering
\includegraphics[width=\textwidth]{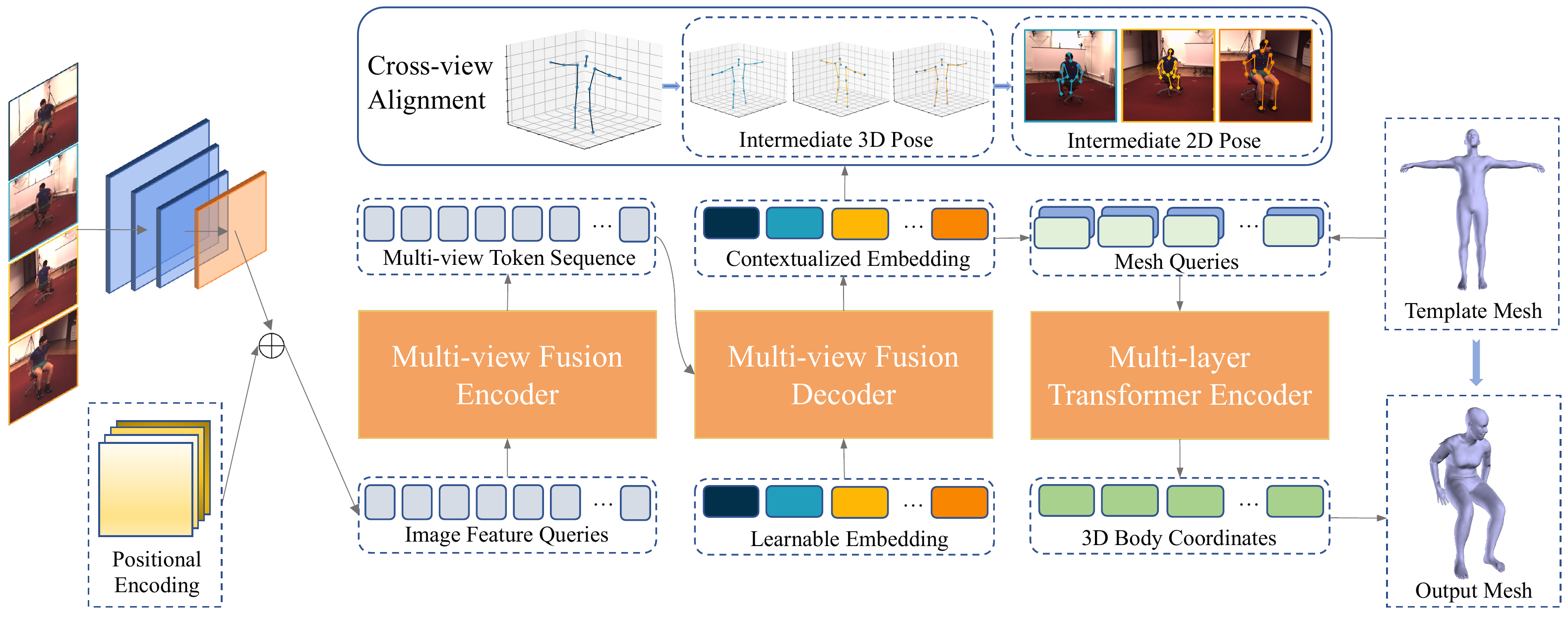}
\vspace{-3mm}
\caption{Overview of the proposed MMT model. Firstly, MMT uses a CNN to encode features for multi-view images. Then, MMT uses encoded features as tokens and feeds them into the multi-view fusion transformer for feature-level fusion to obtain contextualized embedding. To enhance geometry consistency, MMT introduces a cross-view alignment module in the training phase. Finally, MMT utilizes another multi-layer transformer to encode for the targeted human body meshes.
}
\vspace{-3mm}
\label{fig:framework}
\end{figure}

We first briefly introduce the problem and solution setup in a formalized manner. A multi-view non-parametric method for human mesh recovery can be defined as a quad of $\mathcal{(I, V, J, L)}$, where $\mathcal{I}$ denotes the observation space, $\mathcal{(V, J)}$ is the 3D coordinates of mesh vertices and joints respectively, and $\mathcal{L}$ is a loss function of predicted vertices and joints for evaluation and optimization.
Specifically, $\mathcal{I}=\{I_n\}_{n=1}^N$ is a group of images for one person from $N$ diverse viewpoints at the same time, and we set $N=4$ in this paper.
As for the datasets with SMPL annotations, $\mathcal{V}=\{V_n\}_{n=1}^{N} \subset \mathbb{R}^{N \times 6890 \times 3}$ is a set of 6890 vertex coordinates.
Following \cite{johnson2010clustered}, $\mathcal{J}=\{J_n\}_{n=1}^{N} \subset \mathbb{R}^{N \times 14 \times 3}$ contains 3D coordinates of 14 keypoints from different viewpoints to model the skeleton of human body.
In summary, the model takes in RGB images from $N$ perspectives and estimates 3D human pose $\mathcal{J}$ and shape $\mathcal{V}$ under the supervision of ground truth and loss function $\mathcal{L}$. In the following, we will explain the proposed MMT model in detail as shown in Fig.~\ref{fig:framework}.

\subsection{Model Architecture}

\subsubsection{Feature Extract Network}
As the first stage of the proposed method, we use a convolutional image encoder, including ResNet~\cite{he2016deep} and HRNet \cite{wang2020deep}, to obtain global feature vectors $F$ as illustrated below:
\begin{equation}
    F = \mathbf{Concat}(\mathbf{Conv}(I_1), \mathbf{Conv}(I_2), ..., \mathbf{Conv}(I_N)) \in \mathbb{R}^{N \times 7 \times 7 \times 2048}.
\end{equation}
Additionally, the image encoder is pre-trained on ImageNet classification task \cite{russakovsky2015imagenet}, which is beneficial for reconstructing human mesh empirically \cite{lin2021end}.
For each image, we use the vectors from the last layer with the shape of $7\times7\times2048$ and concatenate them for subsequent multi-view feature fusion.

\subsubsection{Multi-view Fusion Transformer}
To construct view-aware features for 3D human pose and shape estimation, it is significant to exert the complementary information embedded among multiple viewpoints.
Different from \cite{carion2020end, zhu2020deformable}, we use transformer to attend to features across images rather than one single image.
Specifically, the multi-view features are processed by Transformer with Multi-Head Attention \cite{vaswani2017attention} and the procedure can be formalized into four steps.
Firstly, the reshaped global feature vector $F$ is fed into a linear projection layer to adjust the dimension:
\begin{equation}
    X = \mathbf{Reshape}(F)W^X = \{x_1, x_2, ..., x_{L}\} \subset \mathbb{R}^{L \times d}.
\end{equation}
Secondly, the input sequence $X$ is projected into the attention triad $(Q, K, V)$ with learned parameters.
Note that the triad of Multi-Head Attention is different from ordinary attention mechanism because it splits the feature representations into $h$ subspaces.
For instance, queries $Q$ is defined as $Q = \{Q_1, Q_2, ..., Q_h\} \subset \mathbb{R}^{h \times L \times \frac{d}{h}}$,
and each element is computed with corresponding projection parameters $(W_i^Q, W_i^K, W_i^V)$ as below:
\begin{equation}
    (Q_i, K_i, V_i) = (XW_i^Q, XW_i^K, XW_i^V).
\end{equation}
Thirdly, we conduct Dot-Product Attention on each subspace and concatenate the results for the context-aware token sequence:
\begin{equation}
\begin{aligned}
    Z = \mathbf{MultiHeadAtt}(Q, K, V) &= \mathbf{Concat}(Y_1, Y_2, ..., Y_h)W^Z \in \mathbb{R}^{L \times d}
\end{aligned} 
\end{equation}
where $Y_i = \mathbf{Softmax}(\frac{Q_iK_i^T}{\sqrt{d/h}})V_i\in \mathbb{R}^{L \times \frac{d}{h}}$.
Finally, the sequence $Z$ is fed into a sublayer to form the final output $\widetilde{Z}$ as follows:
\begin{equation}
    \widetilde{Z} = \mathbf{LayerNorm}(Z + \mathbf{Dropout}(ZW^L)) \in \mathbb{R}^{L \times d}.
\end{equation}
The above algorithm procedure applies to both the transformer encoder and decoder, except for different ways to obtain the query sequence of length $L$.
In the encoder, the length is set as $L_{encoder}=49N$ due to the shape of the global feature vector $F$.
However, to emphasize the semantic information of the human body from different views, we define the length of input queries for the transformer decoder to be $L_{decoder}=KN$, which is inspired by the observation that the human pose contains substantial clues for mesh recovery\cite{choi2020pose2mesh}.
Note that $K$ is the number of keypoints for the skeleton model used by given datasets.
With this design, the final contextualized output sequence $\widetilde{Z}$ not only holds abundant information on the human body, but also contains $N$ token groups attaching to $N$ specific camera views after considering the interactions with other perspectives.

\subsubsection{Cross-view Alignment}
Although multi-view fusion transformer is capable of extracting the underlying relations among different camera views, the generated $\widetilde{Z}$ is not guaranteed to be intensively focused on human pose and shape.
Moreover, $\widetilde{Z}$ is likely to contain conflicting information once the order of input images changes.
Therefore, we propose an auxiliary task, Cross-view Alignment, to mitigate this problem by forcing the model to pay more attention to the human body and global coherence in the observation space.
Specifically, we add a head network to the Multi-view Fusion Transformer to obtain the human pose for all perspectives:
\begin{equation}
    InterPose = \{(p_1^{2D}, p_1^{3D}), (p_2^{2D}, p_2^{3D}), ..., (p_N^{2D}, p_N^{3D})\}.
\end{equation}
Instead of making inferences on $N$ viewpoints each, the head network only outputs the 3D joint locations of one manually-set master view and $N$ groups of camera intrinsics.
With known camera rotation matrix $R_i$, we can obtain the 3D joints for $i$-th view.
Meanwhile, the 3D joints $p_i^{3D}$ can be projected onto 2D surface with $\Pi$ and then transformed with estimated intrinsics $K_i$ for 2D joint locations ${p_i^{2D}}$, which can be formulated as:
\begin{equation}
    p_i^{2D} = \Pi((K_i, R_i), p_i^{3D}) \in \mathbb{R}^{14 \times 2}.
\end{equation}
With this task, $\widetilde{Z}$ is supposed to be more closely related to the human body with cross-view consistency.
Furthermore, making predictions for only one view decreases the computation complexity and empirically helps the head network work more efficiently and precisely.

\subsubsection{Multi-layer Transformer Encoder}
Similar to METRO \cite{lin2021end}, we use a multi-layer transformer encoder with Masked Vertex Modeling to reduce the token dimensions hierarchically and finally map them into 3D keypoints and mesh vertices simultaneously.
Note that when constructing the human body queries, we first use a linear projection layer to make the length of contextualized output sequence $\widetilde{Z}$ and the number of template 3D human joints and vertices match.
Then they are concatenated together to form the input token sequence:
\begin{equation}
\label{query_body}
    Q_{body} = \mathbf{Concat}(\widetilde{Z}W^{body}, \mathbf{Concat}(\mathcal{J}_{T-pose}, \mathcal{V}_{T-pose})).
\end{equation}
For computational simplicity and efficiency \cite{kolotouros2019convolutional}, the $\mathcal{V}_{T-pose}$ is a subset of the all mesh vertices, and it will be upsampled to full size with extra linear projection layers.

\subsection{Training and Inference}
To train the MMT, we construct a loss function by considering both the final inferences and intermediate pose.
Broadly speaking, there are four components within the loss function:
\begin{equation}
    \mathcal{L} = \alpha\mathcal{L_J} + \beta\mathcal{L_V} + \gamma\mathcal{L}_{Align} + \mu\mathcal{L}_{Smooth}.
\end{equation}
Here the $\mathcal{L_J}$ and $\mathcal{L_V}$ are the $L_1$ loss between the final estimation result and the given ground truth.
On the contrary, $\mathcal{L}_{Align}$ is responsible for the optimization of cross-view alignment.
Meanwhile, we apply a smooth term $\mathcal{L}_{Smooth}$ to help the non-parametric model generate more realistic human mesh without any human body prior.

\vspace{-3mm}
\paragraph{Joint and Vertex Loss}
As stated above, the model only needs to make predictions about the master view and then it is feasible to transform the result into other perspectives with camera parameters.
Therefore, $\mathcal{L_J}$, $\mathcal{L_V}$ and $\mathcal{L}_{Smooth}$ only need to be calculated once for the master view between the estimation $(v, j)$ and groud truth $(\hat{v},\hat{j})$, which can be defined as:
\begin{equation}
    \mathcal{L_J} =  \frac{\sum_{i=1}^{K}(\lambda_1|j_{i}^{2D} - \hat{j}_{i}^{2D}| + \lambda_2|j_{i}^{3D} - \hat{j}_{i}^{3D}| +
    \lambda_3|j_{i}^{Reg2D} - \hat{j}_{i}^{2D}| +
    \lambda_4|j_{i}^{Reg3D} - \hat{j}_{i}^{3D}|)}{K}
\end{equation}
where $(j_{i}^{2D}, j_{i}^{3D}) \in J_{master}$.
Besides the direct predictions of $(j_{i}^{2D}, j_{i}^{3D})$, it is worth noting that the $(j_{i}^{Reg2D}, j_{i}^{Reg3D})$ can be obtained from the predicted mesh vertices with a pre-defined regression matrix \cite{kolotouros2019learning, kolotouros2019convolutional, choi2020pose2mesh, kanazawa2018end, lin2021end}.
Furthermore, although the camera intrinsics are accessible, the model still projects the 3D coordinates onto the 2D surface with estimated intrinsics, which helps the model to be more practical and can empirically improve the overall performance \cite{choi2020pose2mesh, lin2021end}.
Similiar to $\mathcal{L_J}$, the loss item on mesh vertices is calculated by:
\begin{equation}
    \mathcal{L_V} =  \eta_1\frac{\sum_{i=1}^{M}|v_{i}^{Full} - \hat{v}_{i}^{Full}|}{M} + \eta_2\frac{\sum_{i=1}^{M_1}|v_{i}^{Sub_1} - \hat{v}_{i}^{Sub_1}|}{M_1} + \eta_3\frac{\sum_{i=1}^{M_2}|v_{i}^{Sub_2} - \hat{v}_{i}^{Sub_2}|}{M_2},
\end{equation}
and $(v_{i}^{Full}, v_{i}^{Sub_1}, v_{i}^{Sub_2}) \in V_{master}$ also exists.
To decrease the time consumption \cite{kolotouros2019convolutional}, the MMT only make inferences on a subset $v_{i}^{Sub_2}$, followed by a two-stage upsampling to progressively recover bigger subset $v_{i}^{Sub_1}$ and full mesh vertices $v_{i}^{Full}$.

\vspace{-3mm}
\paragraph{Cross-view Alignment Loss}
To strengthen the human-related information and geometry consistency in contextualized sequence $\widetilde{Z}$, we use $L_1$ loss to optimize the intermediate output:
\begin{equation}
    \mathcal{L}_{Align} = \sum_{i=1}^{N}\frac{|(p_i^{2D}, p_i^{3D}) - (\hat{J}_i^{2D}, \hat{J}_i^{3D})|}{NK}.
\end{equation}

\vspace{-5mm}
\paragraph{Smooth Loss}
As explained in relevant literature \cite{wang2018pixel2mesh, ge20193d}, applying a smooth loss can help to improve the mesh quality further.
The smooth loss simultaneously attends to normal direction, edge length and the neighboring coherence:
\begin{equation}
    \mathcal{L}_{Smooth} = \sum_{i=1}^{S} \sum_{a, b \in \mathcal{T}(i)}\left\|\left\langle v_{a}-v_{b}, \hat{\mathbf{n}}_{i}\right\rangle\right\|_{2}^{2}+\sum_{i=1}^{E}\left(\left\|\mathbf{e}_{i}\right\|_{2}-\left\|\hat{\mathbf{e}}_{i}\right\|_{2}\right)^{2}+\sum_{i=1}^{M}\left\|\delta_{i}-\frac{\sum_{g \in G_{i}} \delta_{g}}{Card(G_{i})}\right\|_{2}^{2}.
\end{equation}
Specifically, $\langle\cdot,\cdot\rangle$ is the dot product of two vectors and $\hat{\mathbf{n}}_{i}$ is the observed normal of the $i$-th of $S$ pre-defined triangular surfaces.
The second term stands for the square sum of the differences from $E$ edge pairs of estimation and ground truth.
The last term is Laplacian loss for the preserving of local mesh smoothness, where $G_i$ is the set of $Card(G_{i})$ vertices adjacent to $v_i$ and $\delta_{j}=v_j^{3D} -\hat{v}_j^{3D}$ is the offset of the predicted vertex location.

\section{Experiments}
\label{exp}

\subsection{Datasets}
\vspace{-2mm}
\paragraph{HUMBI}
HUMBI~\cite{yu2020humbi} aims to alleviate the problem of data scarcity on recovering human body expressions from multiple camera views.
It provides annotations of 3D joints and body mesh from 107 synchronized HD cameras for 772 distinctive subjects, where 414 are available on the official website. According to the data split ratio (5:2) in Human3.6M, we divide HUMBI into training and testing sets with 294 and 120 subjects respectively. Meanwhile, we select the images of four cameras (14, 32, 50, 67) as the model input for both training and testing. Note that the images in HUMBI usually do not display the whole body of the targeted subject, so it is considered more challenging to recover the body mesh.

\vspace{-3mm}
\paragraph{Human3.6M} 
Human3.6M~\cite{ionescu2013human3} is initially designed for evaluation of 3D human pose estimation, with videos captured from four cameras. Here,
we combine the ground truth 3D pose and pseudo mesh labels \cite{choi2020pose2mesh, moon2020i2l} to train the model.
Following \cite{kanazawa2018end, tome2018rethinking}, we evaluate our method on two standard protocols.
Specifically, P1 requires the model to be trained with all four views of five subjects (S1, S5, S6, S7 and S8) and tested on the other two (S9 and S11).
Although P2 uses the same data split, it only evaluates the model with images taken by camera 3, where the frontal side is the most common.
In other words, P1 is considered more challenging than P2 due to the occurrence of non-frontal images.

\subsection{Implementation Details}
\label{detail}
Following \cite{lin2021end, lin2021mesh}, we use HRNet-W64 \cite{wang2020deep} as the backbone to extract representations of multi-view images.
We implement both the multi-view fusion transformer encoder and decoder with only one layer and eight attention heads.
During training, we use Adam~\cite{kingma2014adam} as the optimizer. We set the initial learning rate as $1\times10^{-4}$, which is decreased by a factor of 0.1 every 100 epochs. The batch size is 32 and the model is trained for 200 epochs in total. We implement the model with PyTorch~\cite{paszke2019pytorch}.

\subsection{Experiments on HUMBI}
To compare proposed method with the state-of-the-art methods, we report the evaluation results in three standard metrics: \textit{Mean Per Joint Position Error} (MPJPE) \cite{ionescu2013human3}, \textit{Procrustes Analysis MPJPE} (PA-MPJPE) \cite{zhou2018monocap} and \textit{Mean Per Vetex Error} (MPVE) \cite{pavlakos2018learning}.

\subsubsection{Comparison with SOTAs}
Due to the absence of baselines on HUMBI, we reproduce two SOTA methods, METRO \cite{lin2021end} and Mesh Graphormer \cite{lin2021mesh} with their public codes for comparison. Following conventions, we use their pre-trained models on 2D and 3D human pose estimation datasets as starting points for finetuning on HUMBI, which are crucial for producing a good performance. Results are shown in Tab.~\ref{humbi_sota}. We can see that MMT greatly outperforms the existing state-of-the-art method \cite{lin2021mesh}, improving the estimation quality by 23.3\%, 25.1\% and 28.8\% on all metrics respectively, even without the usage of any extra datasets. This quantitatively verifies the effectiveness of our MMT model for generating accurate human body meshes. 

\begin{table}[ht]\small
\vspace{-2mm}
\centering
\caption{Comparison with state-of-the-art methods on HUMBI dataset.}
\vspace{-2mm}
\begin{tabular}{lcccc}
\toprule
Method                              & Pre-trained   & MPJPE $\downarrow$    & PA-MPJPE $\downarrow$    & MPVE $\downarrow$  \\ \midrule
METRO \cite{lin2021end}             & Yes           & 50.3                  & 37.7                     & 55.2               \\
Mesh Graphormer \cite{lin2021mesh}  & Yes           & 49.3                  & 36.7                     & 54.8               \\
\textbf{MMT}                        & No            & \textbf{37.8}         & \textbf{27.5}            & \textbf{39.0}      \\ \bottomrule
\end{tabular}
\label{humbi_sota}
\vspace{-2mm}
\end{table}

\subsubsection{Ablation Studies}

\paragraph{Studies on cross-view alignment}
The auxiliary task allows the multi-view fusion transformer to insert 3D human body information into the fusion process and ensure the semantic tokens contain coherent clues across different views.
According to Tab.~\ref{humbi_cross_view}, we can find that only supervising 3D pose is not enough, and it should be combined with 2D pose alignment, i.e., supervising both 3D pose and camera intrinsics helps the model learn better.
Furthermore, we attempt to replace the T-pose template in Eq.~\ref{query_body} with the intermediate pose estimations for multi-layer transformer encoder.
The result proves again that the coarse-to-fine manner is capable of reconstructing the human body from multi-view features. 
Thus we need not import other body-related information into the refining process.

\begin{table}[ht]\small
\vspace{-2mm}
\caption{Ablation studies on the cross-view alignment module experiments on HUMBI dataset.}
\vspace{-2mm}
\centering
\begin{threeparttable}
\begin{tabular}{cccc}
\toprule
Intermediate Task                       & MPJPE $\downarrow$ & PA-MPJPE $\downarrow$  & MPVE $\downarrow$ \\ \midrule
—                                       & 41.4               & 30.7                   & 43.7              \\
3D Alignment                            & 41.7               & 29.5                   & 44.7              \\
3D\&2D Alignment                        & \textbf{37.8}      & \textbf{27.5}          & \textbf{39.0}     \\
3D\&2D Alignment + Template Replacement & 38.9               & 28.1                   & 40.3              \\ \bottomrule
\end{tabular}
\small
\end{threeparttable}
\label{humbi_cross_view}
\vspace{-2mm}
\end{table}

\vspace{-3mm}
\paragraph{Studies on fusion strategies}
We implement three other feature-level fusion strategies to illustrate the effectiveness of 
the carefully designed architecture for multi-view fusion in MMT.
The first MMT-$1\times1$ Convolution uses a $1 \times 1$ convolution to merge the feature maps from different cameras into a single one.
The second MMT-{Fusion Strategy A}
takes the feature tokens and coarse predictions from the previous stage and then outputs refined estimation results with only supervision on a 3D pose for the master view.
The third MMT-Fusion Strategy B performs output-level fusion by transforming the estimation result into three other views and matching them with the corresponding ground truth labels.~\footnote{More details about alternative fusion strategies can be found in the supplementary material.} Results are shown in Tab.~\ref{humbi_feat_fuse}. We can see that simple fusion with $1\times1$ convolution cannot capture the multi-view priors, resulting in poor results. By comparing MMT with its variant MMT-Fusion Strategy A, we can find multi-view supervision can help to produce more accurate results. This shows the effects of our fusion strategy to capture multi-view priors globally. By comparing MMT with its variant MMT-Fusion Strategy B, we can find that the proposed feature-level fusion achieves notable improvement over existing output-level fusion.

\vspace{-3mm}
\begin{table}[ht]\small
\centering
\caption{Ablation studies on the feature-level fusion strategies on HUMBI dataset.}
\vspace{-2mm}
\begin{tabular}{lcccc}
\toprule
Method                      & Pre-trained   & MPJPE $\downarrow$    & PA-MPJPE $\downarrow$    & MPVE $\downarrow$  \\ \midrule
MMT-$1\times1$ Convolution  & No            & 98.9                  & 86.8                     & 135.3              \\
MMT-Fusion Strategy A       & No            & 40.3                  & 29.9                     & 46.7               \\
MMT-Fusion Strategy B       & No            & 40.4                  & 30.1                     & 43.4               \\
\textbf{MMT}                & No            & \textbf{37.8}         & \textbf{27.5}            & \textbf{39.0}      \\ \bottomrule
\end{tabular}
\label{humbi_feat_fuse}
\end{table}

\vspace{-3mm}
\paragraph{Studies on the number of camera views}
It is clear that cameras from different directions contain massive complementary information to reconstruct the target object, which allows the model to handle challenging conditions like the incomplete and occluded human body.
Therefore, we deploy MMT under various camera settings and observe its reactions.
We can tell from Tab.~\ref{humbi_no_sl} that with the increasing camera views, our proposed method can generate more accurate human pose and shape accordingly.
The results mean that MMT can effectively fuse and leverage the multi-view clues for human mesh recovery as expected.

\vspace{-3mm}
\paragraph{Studies on the importance of Smooth Loss}
According to the visualization result in the \textit{supplementary material}, it is clear that the import of $\mathcal{L}_{Smooth}$ helps the non-parametric model make more realistic inferences about the human body.
However, similar to the scene reconstruction \cite{wang2018pixel2mesh}, we discover that existing metrics on the human mesh cannot provide a complete picture of the model's ability.
Specifically, as Tab.~\ref{humbi_no_sl} shows, the model trained with $\mathcal{L}_{Smooth}$ can generate human mesh with fewer bulges but fails to outperform the original model on quantitative metrics.

\textbf{More ablation studies on transformer architectures are provided in the supplementary material.}

\begin{table}[t!]\small
\caption{Ablation studies on effects of view number and smooth loss on HUMBI dataset.}
\begin{minipage}{0.5\linewidth}
\begin{threeparttable}
\centering
\setlength\tabcolsep{3.75pt}
\begin{tabular}{cccc}
\toprule
View Number & MPJPE $\downarrow$    & PA-MPJPE $\downarrow$    & MPVE $\downarrow$  \\ \midrule
1           & 59.3                  & 43.2                     & 66.9               \\
2           & 47.8                  & 33.9                     & 52.0               \\
3           & 42.2                  & 31.6                     & 44.9               \\
4           & \textbf{37.8}         & \textbf{27.5}            & \textbf{39.0}      \\ \bottomrule    
\end{tabular}
\small
\end{threeparttable}
\end{minipage}
\begin{minipage}{0.5\linewidth}
\begin{threeparttable}
\centering
\setlength\tabcolsep{3.6pt}
\begin{tabular}{cccc}
\toprule
$\mathcal{L}_{Smooth}$  & MPJPE $\downarrow$    & PA-MPJPE $\downarrow$    & MPVE $\downarrow$     \\ \midrule
No                      & 37.8                  & \textbf{27.5}            & \textbf{39.0}         \\
Yes                     & \textbf{37.4}         & 27.6                     & 43.3                  \\ \bottomrule 
\end{tabular}
\small
\end{threeparttable}
\end{minipage}
\label{humbi_no_sl}
\vspace{-5mm}
\end{table}

\subsection{Experiments on Human3.6M}
Firstly, we evaluate our proposed method on Human3.6M under two standard protocols and compare its performance with the state-of-the-art methods.
As Tab.~\ref{h36m_sota} shows, the MMT outperforms all previous methods under P1, reducing MPJPE by 11.2\% than the best monocular method \cite{lin2021mesh} without extensive pre-training datasets \cite{moon2020i2l, omran2018neural, kanazawa2018end, kolotouros2019convolutional, choi2020pose2mesh} and delicate data augmentation \cite{lin2021end}.

\begin{table}[h!]\footnotesize
\vspace{-1mm}
\caption{Comparison with state-of-the-art methods on Human3.6M dataset with protocol 1 and 2.}
\vspace{-2mm}
\begin{minipage}{0.5\linewidth}
\setlength\tabcolsep{2pt}
\begin{threeparttable}
\centering
\begin{tabular}{lcc}
\toprule
\multirow{2}{*}{Method}  & \multicolumn{2}{c}{Protocol 1}  \\ \cline{2-3} 
                         & MPJPE $\downarrow$      & PA-MPJPE $\downarrow$    \\ \midrule
\multicolumn{3}{c}{Single-view Methods} \\
DenseRac \cite{xu2019denserac}                 & 76.8           & -              \\
HKMR \cite{georgakis2020hierarchical}                     & 64.0           & -              \\
HMR \cite{kanazawa2018end}                     & 87.9           & 58.1           \\
GraphCMR \cite{kolotouros2019convolutional}                & 74.7           & 51.9           \\
Martinez \textit{et al.} \cite{martinez2017simple}        & 62.9           & 47.7           \\
DecoMR \cite{zeng20203d}                   & 62.7           & 42.2           \\
METRO \cite{lin2021end}                   & 60.2           & 42.2           \\
Mesh Graphormer \cite{lin2021mesh}         & 56.5           & 39.7           \\ \hline \hline
\multicolumn{3}{c}{Multi-view Methods} \\
Trumble \textit{et al.} \cite{trumble2018deep}          & 62.5           & -              \\
Tome \textit{et al.} \cite{tome2018rethinking}             & 52.8           & -              \\
Liang \textit{et al.} \cite{liang2019shape}             & 79.8           & 45.1           \\
Li \textit{et al.} \cite{li20213d}                      & 64.8           & 43.8           \\
\textbf{MMT} & \textbf{50.2}  & \textbf{37.3}  \\ \bottomrule
\end{tabular}
\small
\end{threeparttable}
\end{minipage}
\begin{minipage}{0.5\linewidth}
\setlength\tabcolsep{2pt}
\begin{threeparttable}
\centering
\begin{tabular}{lcc}
\toprule
\multirow{2}{*}{Method}  & \multicolumn{2}{c}{Protocol 2} \\ \cline{2-3} 
                         & MPJPE $\downarrow$      & PA-MPJPE $\downarrow$    \\ \midrule
\multicolumn{3}{c}{Single-view Methods} \\
HKMR \cite{georgakis2020hierarchical}                     & 59.6           & -             \\
HMR \cite{kanazawa2018end}                      & -              & 56.8          \\
GraphCMR \cite{kolotouros2019convolutional}                 & 71.9           & 50.1          \\
Pose2Mesh \cite{choi2020pose2mesh}               & 64.9           & 47.0          \\
HoloPose \cite{guler2019holopose}                & 60.2           & 46.5          \\
VIBE \cite{kocabas2020vibe}                    & 65.6           & 41.4          \\
SPIN \cite{kolotouros2019learning}                    & -              & 41.1          \\
I2L-MeshNet \cite{moon2020i2l}             & 55.7           & 41.1          \\
DecoMR \cite{zeng20203d}                   & 60.6           & 39.3          \\
METRO \cite{lin2021end}                   & 54.0           & 36.7          \\
Mesh Graphormer \cite{lin2021mesh}         & 51.2  & \textbf{34.5} \\ \hline \hline
\multicolumn{3}{c}{Multi-view Methods} \\
Tome \textit{et al.} \cite{tome2018rethinking}              & -              & 47.6          \\
\textbf{MMT} & \textbf{50.2}           & 37.1          \\ \bottomrule
\end{tabular}
\small
\end{threeparttable}
\end{minipage}
\label{h36m_sota}
\vspace{-5mm}
\end{table}

Although the multi-camera approach with restrictions on geometry consistency \cite{tome2018rethinking} can generate good results on MPJPE as our method, it depends on intermediately estimated 2D keypoints and cannot reconstruct human body mesh.
We can tell from Tab.~\ref{h36m_sota} that our model achieves quite competitive results under P2, which favors the single-view methods greatly.
However, it is interesting to find that even under P1, the single-view method \cite{lin2021mesh} is clearly superior to multi-view methods \cite{trumble2018deep,liang2019shape, li20213d} on both metrics, not to mention the results under P2.
Additionally, benefiting from the context-aware tokens, MMT is allowed to achieve excellent performance under both protocols regardless of the views to make inferences about, which is not seen in previous state-of-the-art methods \cite{lin2021mesh, liang2019shape, tome2018rethinking}.

\subsection{Qualitative Results} 
\begin{figure}[t!]
\centering
\includegraphics[width=0.9\textwidth]{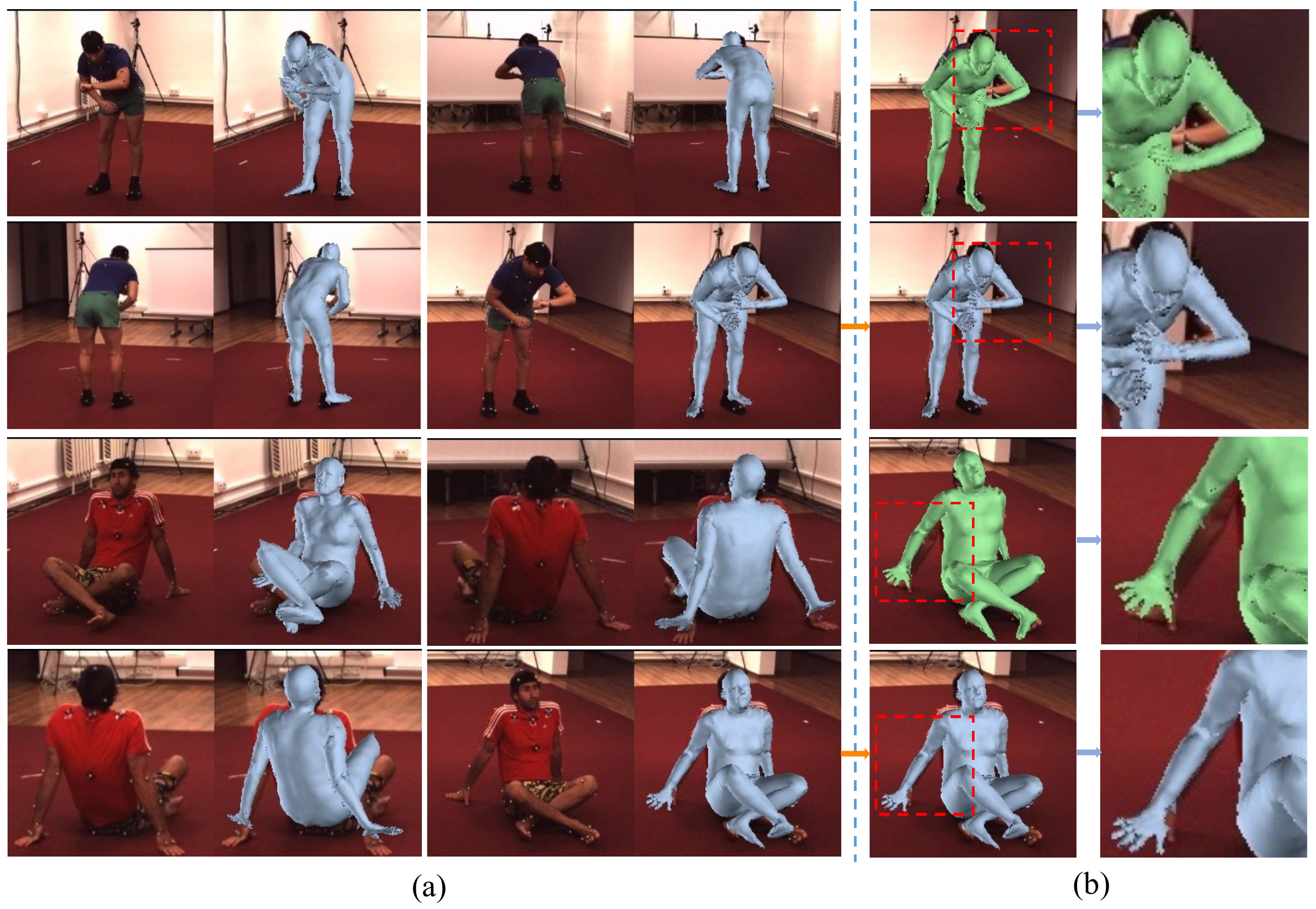}
\vspace{-3mm}
\caption{Qualitative results. (a) Visualization results of MMT on Human3.6M. Every subject has a group of four multi-view images paired with corresponding reconstruction results from MMT, and each group takes up two rows. (b) Comparsion between MMT (\textcolor[rgb]{0.71372549, 0.78823529, 0.91372549}{blue}) and Mesh Graphormer (\textcolor[rgb]{0.67843137, 0.89019608, 0.62352941}{green}). Best viewed in color.
}
\label{fig:vis_rlt}
\vspace{-5mm}
\end{figure}

We presents qualitative results in Fig.~\ref{fig:vis_rlt}. From Fig.~\ref{fig:vis_rlt} (a), we can see that MMT can recover accurate and consistent body meshes of different views, with pose variation and occlusions. From Fig.~\ref{fig:vis_rlt} (b), we can find that MMT can produce estimations that fit the subjects better over the single-view SOTA Mesh Graphormer. This further demonstrates that the proposed MMT model can leverage the multi-view guidance to overcome the ill-posed problem occurring to the single-view counterparts. \textit{More visualization results can be found in the supplementary material.}

\section{Conclusion}
\label{conclusion}
In this paper, we propose a novel multi-view framework, Multi-view human body Mesh Translator, to reconstruct 3D human pose and shape from multi-view images. Different from existing methods with output-level fusion, MMT conducts the feature-level fusion to leverage the multi-view guidance more sufficiently, thus deriving improved mesh recovery results. In addition, MMT presents a new cross-view alignment strategy, which introduces geometry constraints at the feature level and encourages estimation consistency for different views. Comprehensive experiments on multiple benchmarks demonstrate the effectiveness of the proposed MMT model for "translating" view-wise images to their corresponding targeted meshes. 

During the verification stage, we find that the existing evaluation metrics are still rough and thus fail to denote the actual reconstruction ability of the model.
In the future, we will study in this direction and try to put forward more inclusive standards to measure the quality of estimated 3D human pose and shape.
Furthermore, using the non-parametric model for human mesh recovery is inevitably a double-edged weapon.
It is easier to adopt MMT for other tasks than model-based methods, like recovering mesh for hands or other objects.
However, the speed of model-free algorithms is slower than the parametric ones.
Therefore, we will also look into this limitation and extend MMT to a more practical tool in daily life.

\bibliographystyle{plain}
\bibliography{neurips_2022}

\end{document}